\pdfoutput=1

\documentclass[11pt]{article}

\usepackage{acl}

\usepackage{times}
\usepackage{latexsym}

\usepackage[T1]{fontenc}

\usepackage[utf8]{inputenc}

\usepackage{microtype}

\usepackage{inconsolata}

\usepackage{booktabs}
\usepackage{multirow}

\usepackage{graphics}
\usepackage{graphicx}

\title{German Text Simplification: Finetuning Large Language Models with Semi-Synthetic Data}

\author{
    Lars Klöser \and Mika Beele \and Jan-Niklas Schagen \and Bodo Kraft \\
    University of Applied Sciences Aachen \\
    \texttt{\{kloeser, beele, kraft\}@fh-aachen.de} \\
    \texttt{niklas.schagen@alumni.fh-aachen.de}
}
\begin{document}
\maketitle
\begin{abstract}
This study pioneers the use of synthetically generated data for training generative models in document-level text simplification of German texts. We demonstrate the effectiveness of our approach with real-world online texts. Addressing the challenge of data scarcity in language simplification, we crawled professionally simplified German texts and synthesized a corpus using GPT-4. We finetune \textit{Large Language Models} with up to 13 billion parameters on this data and evaluate their performance. This paper employs various methodologies for evaluation and demonstrates the limitations of currently used rule-based metrics. Both automatic and manual evaluations reveal that our models can significantly simplify real-world online texts, indicating the potential of synthetic data in improving text simplification.
\end{abstract}

\section{Introduction}

In our modern and digitalized societies, access to information is essential for active participation. However, certain groups, such as individuals with intellectual disabilities or non-native speakers, often struggle to understand the local language, which can impede their social and civic engagement. Each group faces unique challenges in text comprehension. Integrating automatic text simplification tools can significantly benefit these groups by providing accessible information, thereby providing a pivotal means for greater inclusion.

Various linguistic initiatives, like \cite{netzwerk_leichte_sprache_regeln_2024}, have been established in German-speaking regions to address this, specifically designed for different target groups. Such efforts align with international legal frameworks like \textit{Article 9 of the UN Convention on the Rights of Persons with Disabilities}\footnote{\url{https://www.ohchr.org/en/instruments-mechanisms/instruments/convention-rights-persons-disabilities}}, which advocates for the right to accessible communication.

Creating simplified content manually is a labor-intensive and time-consuming process, significantly hindering its broad availability and accessibility. In contrast, \textit{Large Language Models} (LLMs), especially smaller ones fine-tuned for text simplification, offer a viable and efficient alternative \cite{anschutz_language_2023}. These smaller models require fewer resources and are simpler to operate than larger LLMs, making them ideal for scaling up the process of automatic language simplification. A key challenge in finetuning LLMs for text simplification lies in the limited availability of parallel data \cite{anschutz_language_2023, toborek_new_2022}.

\begin{figure}
    \centering
    \includegraphics[scale=.1]{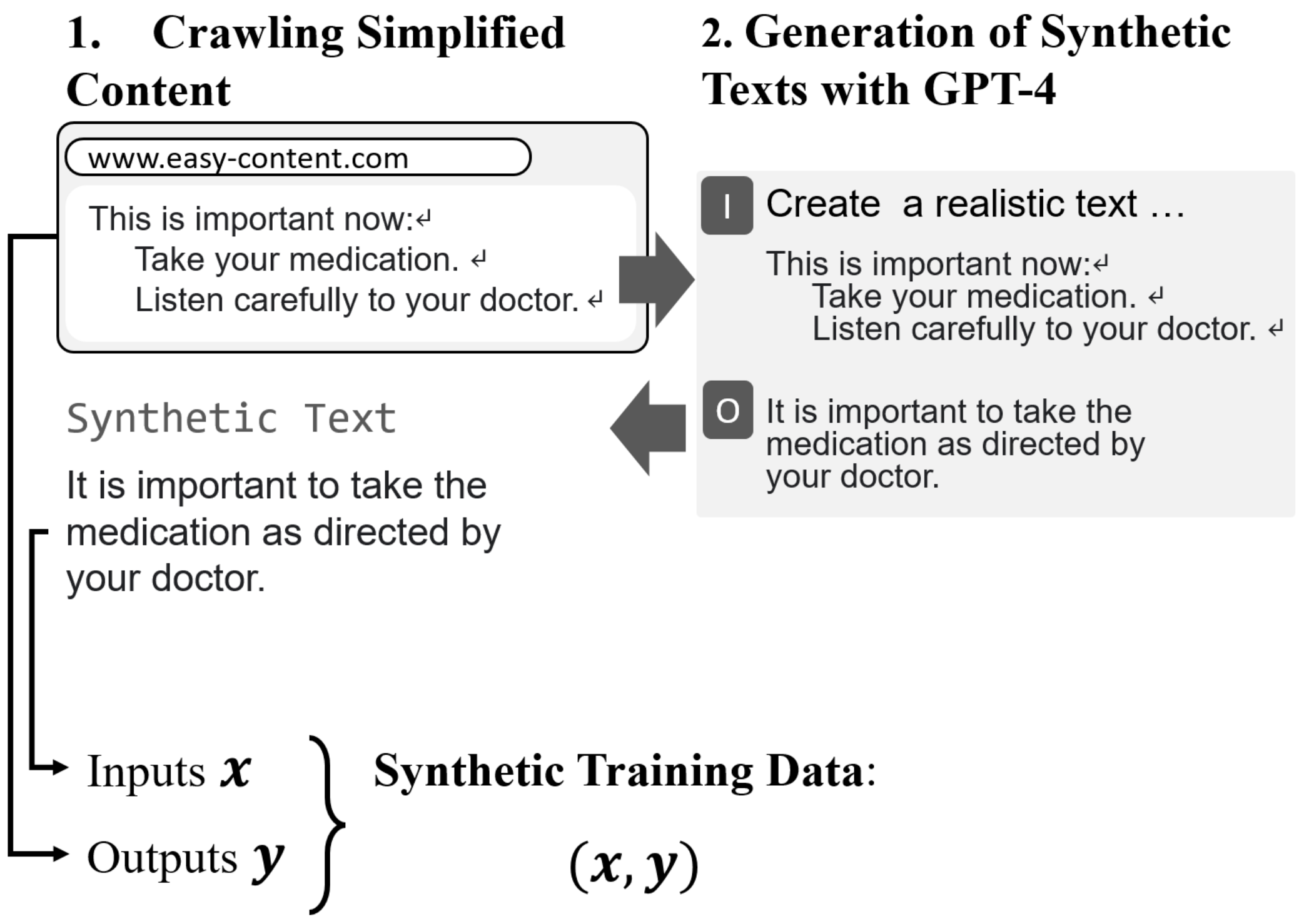}
    \caption{Illustration of synthetic data generation. Data is crawled from websites specializing in language simplification. GPT-4 generates texts in everyday language, ensuring the original content remains unaltered. We construct a simplification dataset where these texts serve as input while the crawled simplifications act as reference simplifications.}
    \label{fig:concept}
\end{figure}

Our approach, as illustrated in \autoref{fig:concept}, tackles the challenge of data scarcity in language simplification by creating semi-synthetic data. This involves crawling various sources for already simplified web content and then utilizing GPT-4 to generate hypothetical original texts corresponding to these simplifications. Our dataset thus comprises GPT-4's outputs as the inputs and the crawled content as the simplified outputs, forming a text simplification dataset. We apply this dataset as the basis for finetuning Large Language Models (LLMs) for automatic text simplification.

We publish all necessary resources to reproduce this paper's results on a public GitHub repository\footnote{\url{https://github.com/MSLars/German-Text-Simplification}}. Our scientific contributions can be summarized as follows:
\begin{enumerate}
    \item Creating a corpus of parallel text simplification data in German based on novel methodology.
    \item Training, evaluating, and releasing LLM-based language simplification models for German texts.
\end{enumerate}

\section{Related Work}

Various approaches and methodologies have been developed for automatic German text simplification. \cite{anschutz_language_2023} proposed a two-step approach utilizing pretrained language models finetuned on German simplifications to diminish the requirement for parallel data. In contrast to our approach, the parallel data contains a mixture of summarization and simplification and targets only newspaper articles. \cite{spring_exploring_2021} train German text simplification models by using labels to target specific language levels, ensuring model adaptations are level-appropriate and control copying behavior. Their dataset focuses on newspaper articles and sentence-level simplification.

Diverging from neural network-based methods, \cite{garain_sentence_2019} introduced methodology based on parse trees. Similarly, \cite{praveen_kumar_pattern-based_2022} offered a pattern-based syntactic simplification framework. \cite{kajiwara_text_2018} presented methods for text simplification in languages with limited simplified corpora, including lexical substitution and monolingual translation, focusing on resource-scarce languages like Japanese.

Regarding evaluation metrics, \cite{sulem_bleu_2018} investigate the limitations of BLEU as a widespread evaluation metric for text generation tasks. \cite{alva-manchego_suitability_2021} explored the correlation between existing metrics and human judgments in multi-operation text simplifications, providing insights into the appropriateness of automatic metrics for assessing text simplification. \cite{maddela_lens_2022} introduced LENS, a learnable evaluation metric for text simplification trained on modern language models, showing a better correlation with human judgment.

Regarding parallel corpora and resources, \cite{ebling_automatic_2022} aggregated corpora for the automatic processing of simplified German, providing resources for training and evaluation. \cite{holmer_constructing_2023} create so-called \textit{pseudo parallel} sentence pairs of simple and complex sentences from given sentence collections. \cite{hauser_multilingual_2022} introduced SNIML, a multilingual corpus of news articles in simplified language, and \cite{rios_new_2021} showcased a dataset for document-level text simplification in German, including articles paired with simplified summaries. \cite{aumiller_klexikon_2022} addressed the challenge of concurrently summarizing and simplifying longer texts, introducing a new dataset for joint text simplification and summarization. \cite{hewett_apa-rst_2023} introduces a dataset with parallel sentence-level simplications and additional information about the document's rhetorical structure.

Text simplification corpora exist for various languages. For example, \cite{coster_simple_2011} introduced a dataset pairing English Wikipedia with Simple English Wikipedia, enabling the analysis of various simplification operations, including rewording, reordering, insertion, and deletion.

In Easy Language generation, \cite{deilen_using_2023} investigated the feasibility of using ChatGPT to translate administrative texts into German \textit{Leichte Sprache} (easy language), a highly regulated language variety with a focus on text simplification.

\section{Task Definition}
\label{sec:task_definition}

In this section, we give a task definition for language simplification. This definition outlines the requirements for the trained models and motivates our methodology for dataset creation.

The \textbf{inputs space} consists of editorially created German texts. Based on the selection of web sources, we assume predominantly grammatically complete sentences and quality-assured content. In the context of this work, we exclude user-created input and social media texts. These web documents contain multiple paragraphs and sentences.

The \textbf{output space} consists of simplifications of the input. The style and level of simplification correspond to the contents currently available in the German language, precisely as they are presently accessible. We aim to avoid modifications of the content, like summarization. However, understandability may require additional explanations of certain concepts in the simplified texts.

Two concepts for language simplification have been established for the German language: \textit{Einfache Sprache} (simple language) and \textit{Leichte Sprache} (easy language). We seek to explain how our methodologies intersect and align with these well-established frameworks.

Simple language covers text simplification in general. Possible target groups include readers unfamiliar with the domain or used language, for example, in legal or medical texts or language learners. The target group can fundamentally understand the concepts. Linguistic complexity, however, makes understanding more difficult. Moreover, these simplifications can aid in language acquisition. 

In contrast, easy language is aimed at people with severely limited text comprehension, such as those with intellectual disabilities. Fixed sets of simplification rules have been established \cite{netzwerk_leichte_sprache_regeln_2024}. These rules cover areas like syntactical, lexical, or typographical simplifications.

In our approach, we scrape texts from various sources, each characterized by its distinct language style. Our research hints that these target texts generally do not conform to the rules of easy language. Rather, many crawled texts may align more with the domain of simple language.

\subsection{Dataset}
\label{taskdefinition:dataset}

\begin{table}[]
\caption{Word and document frequencies in the dataset across different sources, segregated into test and train sets}
\centering
\scriptsize
\begin{tabular}{lrrrr}
\toprule
                           & \multicolumn{2}{c}{Test} & \multicolumn{2}{c}{Train} \\
                           & Docs & Words & Docs & Words \\
\midrule
\textit{einfachstars}      & 317       & 45,444     & 2,213      & 307,307 \\
\textit{mdr}               & 10        & 1,696     & 85        & 13,285 \\
\textit{nachrichtenleicht} & 298       & 44,138     & 2,147      & 318,069 \\
\textit{hurraki}           & 181       & 16,152     & 1,234      & 109,386 \\
\textit{ndr}               & 94        & 17,211     & 709       & 135,340 \\
\textit{kurier}            & 72        & 13,425     & 481       & 76,744 \\
\textit{leicht-kicken}     & 8         & 537     & 67        & 2,672 \\
\textit{einfach-teilhaben} & 8         & 749     & 79        & 8,193 \\
\textit{stadt-koeln}       & 3         & 2,588     & 16        & 11,830 \\
\textit{inclusion\_europe} & 1         & 35     & 18        & 920 \\
\textit{bundesregierung}   & 5         & 1,337     & 22        & 9,212 \\
\textit{hamburg-de}        & 3         & 1,199     & 59        & 16,280 \\
\midrule
$\sum$                     & 1000      & 144,511     & 7,130      & 1,009,238 \\
\bottomrule
\end{tabular}
\label{tab:data_set_size}
\end{table}

We introduce a parallel corpus consisting of texts in everyday language and their corresponding simplifications as an instantiation of the task defined in \autoref{sec:task_definition}.

We create semi-synthetical text pairs to overcome the challenge of training data scarcity. Based on the results of various benchmarks, we assume that GPT-4 can produce human-like texts in various domains \cite{openai_gpt-4_2023}. We crawl simplified texts by expanded versions of the crawlers used by \cite{anschutz_language_2023}. Our additional preprocessing standardizes the typography using rule-based methods. Subsequently, we use GPT-4 to create realistic synthetic source texts from the simplifications. To ensure the generated texts were sufficiently diverse, 15 distinct prompts were used. In the following, we will investigate these data in detail.

\subsubsection{Synthetic Texts}

\autoref{tab:data_set_size} presents the scope and size of the semi-synthetic dataset created for German text simplification. A random sample of 1,000 examples has been reserved as test data. To our knowledge, this dataset is the first semi-synthetic approach to the German language simplification task. It is also noteworthy for being the most comprehensive dataset available for document-wide simplification and the only dataset focusing on document-level language simplification across various domains.

The performance of machine learning models is highly contingent on the quality of the training data, as indicated in various studies \cite{jain_overview_2020}. We believe that the complexity and characteristics of the synthetic data used for training should closely mirror the real data from similar contexts in the specific field.

\begin{table}[ht]
\caption{Analysis of textual complexity across various domains. Synthetic web content is slightly more complex compared to real web content. The metrics support that the crawled simplifications are less complex than real and synthetic everyday web content.} 
\centering
\scriptsize 
\begin{tabular}{p{.55cm}ccc}
\toprule
\addlinespace[0.5ex]
& Sports & Celebrities & News \\
\midrule
\multicolumn{4}{l}{\textit{Metric: avg. sentence length}} \\
\addlinespace[0.25ex]
Easy  & $10.38 \pm 2$ & $11.56 \pm 2.45$ & $10.79 \pm 1.49$ \\
Synth.& $24 \pm 9.3$ & $22.94 \pm 6.59$ & $21 \pm 5.4$ \\
Com.  & $19.59 \pm 3.65$ & $21.17\pm 4.21$ & $18.66 \pm 2.7$ \\
\addlinespace 
\midrule
\multicolumn{4}{l}{\textit{Metric: avg. commas per sentence}} \\
\addlinespace[0.25ex]
Easy  & $.09 \pm .18$   & $.17 \pm .24$ & $.00 \pm .03$ \\
Synth.& $1.69 \pm 1.18$  & $1.67 \pm .81$ & $1.52 \pm .72$ \\
Com.  & $.48 \pm .41$ & $1.3 \pm .52$ & $.81 \pm .35$ \\
\addlinespace 
\midrule
\multicolumn{4}{l}{\textit{Metric: avg. distance verb compounds}} \\
\addlinespace[0.25ex]
Easy  & $.09 \pm .15$ & $.14 \pm .14$ & $.14 \pm .11$ \\
Synth.& $.34 \pm .33$ & $.36 \pm .19$ & $.34 \pm .19$ \\
Com.  & $.27 \pm .19$ & $.3 \pm .14$  & $.3 \pm .13$ \\
\bottomrule
\end{tabular}
\label{tab:linguistic_complexity_synthetic}
\end{table}

\autoref{tab:linguistic_complexity_synthetic} offers an overview of various metrics used to estimate the linguistic complexity of the crawled simplified texts, the synthetic data, and real German web content, examining three distinct domains as examples. The selected metrics rate the reconstructed texts slightly more complex than the crawled texts. While these metrics do not definitively determine whether the data is realistic and overcomplication in reconstruction cannot be ruled out, our preliminary conclusion is that the examples could be suitable for the task.

Given that the primary focus of this work is not on the realistic generation of web content, we do not delve deeper into these aspects. Instead, our research examines whether the trained models effectively reduce the complexity of real web content as evaluated in \autoref{sec:real_word_evaluation}. This approach aligns with our goal to enhance the practical applicability of language simplification tools in real-world scenarios.

\subsection{Crawled Simplifications}

This section delves into the specifics of the crawled simplifications. \autoref{tab:linguistic_complexity_synthetic} categorizes various sources into domains. This offers a structured view of the different simplifications obtained from these domains.

\begin{table}[ht]
\caption{Comparative analysis between different  styles of simplified news content.}
\centering
\scriptsize 
\begin{tabular}{p{1.35cm}ccc}
\toprule
\addlinespace[0.5ex]
\textit{Metrik} & MDR & NDR & NL \\
\midrule
\multirow{2}{=}{\textit{Sentence length}} & \multirow{2}{*}{$12.39 \pm 1.54$} & \multirow{2}{*}{$10.47 \pm 1.43$} & \multirow{2}{*}{$12 \pm 1.5$} \\
& & & \\
\midrule
\multirow{2}{=}{\textit{Commas per sentence}} & \multirow{2}{*}{$.04 \pm .09$} & \multirow{2}{*}{$.00 \pm .00$} & \multirow{2}{*}{$.22 \pm .17$} \\
& & & \\
\midrule
\multirow{2}{=}{\textit{Distance verb compounds}} & \multirow{2}{*}{$.2 \pm .12$} & \multirow{2}{*}{$.14 \pm .11$} & \multirow{2}{*}{$.21 \pm .14$} \\
& & & \\
\midrule
\multirow{2}{=}{\textit{Words per line}} & \multirow{2}{*}{$8.78 \pm 22.12$} & \multirow{2}{*}{$8.55 \pm 11.77$} & \multirow{2}{*}{$14.14 \pm 24.15$} \\
& & & \\
\bottomrule
\end{tabular}
\label{tab:ndr_mdr_nachrichten_leicht_compare}
\end{table}

\autoref{tab:ndr_mdr_nachrichten_leicht_compare} investigates the variety inside a single domain. It comprehensively analyzes metrics related to simplified texts from various news providers. These metrics reveal notable differences in the style of simplifications among the providers. For instance, NDR's texts stand out for their absence of commas, suggesting a preference for simpler sentence structures without subordinate clauses. In contrast, NL (nachrichtenleicht.de) frequently employs commas, indicating a higher likelihood of compound and complex sentences, often incorporating subordinate clauses. Additionally, NL's texts, on average, contain longer sentences than other sources, highlighting a distinct approach to text simplification. These findings underscore the stylistic diversity within the dataset, demonstrating that simplifications are not uniform but vary significantly across news providers.

\section{Methodology}

This study aims to perform task-specific fine-tuning of LLMs. The extensive volume of publications in this domain makes it impractical to evaluate all available models and training configurations against one another. Instead, we aim to justify our core design choices in this chapter and provide further justification through targeted evaluation in the subsequent chapter.

\subsection{Language Modelling}

We apply LLMs based on so-called \textit{decoder-only} transformer models as introduced in \cite{radford_improving_2018}. Decoder-only models are designed to model the probability of the subsequent token $P(x_{i+1} | x_1,\dots,x_{i})$ in a given sequence 
$x=x_1,\dots,x_n$ of tokens that represent a text. We represent each text simplification sample as a sequence $x=(x_{source}, SEP, x_{target})$ where $SEP$ is a special token that separates the source from its simplification target. As input, we provide $x_{source}$ followed by the SEP token. The model, during training, attempts to maximize the probabilities

$$
P(x^t_{i+1}|x_{source}, x^t_1,\dots,x^t_i)\;,i=1,\dots,m
$$

of the tokens in the simplifications $x_{target} = x^t_1,\dots,x^t_m$ using the cross entropy loss.

We finetuned two distinct versions of two different pretrained LLMs. Specifically, we finetuned two German versions of GPT-2 \cite{minixhofer_wechsel_2022} and two versions of the Leo LM model \cite{pluester_leolm_2023}. We selected these models because they are decoder-only, available in different sizes and pretrained on german texts. Each of these models underwent finetuning on our training dataset for three epochs. For this process, we employed the HuggingFace\footnote{https://huggingface.co/} library, a popular choice for machine learning and natural language processing tasks. The detailed configurations used for training, including parameters and environmental settings, are meticulously documented in \autoref{sec:appendix}. 

\subsection{Decoding Algorithm}
\label{sec:decoding_algo}
This section describes the methods of deriving concrete sequences from the probability distributions for individual follow-up tokens provided by LLMs, commonly called \textit{decoding algorithms}. We compare four distinct approaches:

\textbf{Greedy Approach}: This method sequentially selects the token with the highest probability. It is straightforward and efficient but may not yield the most contextually appropriate sequence.

\textbf{Beam Search Algorithm}: This technique chooses the best alternative from a fixed number of possibilities, each with the currently highest probability. It balances between exploring various possibilities and focusing on the most probable options.

\textbf{Sampling-Based Algorithm}: Here, follow-up tokens are randomly selected based on the probability distribution of the LLMs. This approach introduces variability and can generate more diverse outputs \cite{holtzman_curious_2020}.

\textbf{Contrastive Search Approach}: This novel method contrasts traditional search techniques. It considers the likelihood of individual tokens and evaluates the probability distribution over a set of potential sequences, aiming to balance between the most probable and contextually appropriate choices. This approach is useful in ensuring that the generated text maintains coherence and relevance \cite{su_contrastive_2022}.

We utilized a fixed configuration for each approach as provided in \autoref{sec:appendix}. This comparative analysis offers insights into the efficacy and suitability of different decoding strategies.

We frequently observed prediction repetitions in our investigation, particularly with smaller models.
In text generation, in general, a repetition penalty is frequently used. However, in this context, some repetition may be beneficial. Hence, we've devised an alternative approach that allows for a certain degree of repetition, recognizing its potential value in making texts clearer and more comprehensible.

To address this, we implemented a strategy to halt the generation of further tokens if the frequency of a token within a certain window exceeded a predefined threshold. This intervention was designed to enhance the quality of the generated text by preventing excessive repetition, which can detract from the readability and coherence of the output. Such a method is crucial in maintaining language's natural flow and diversity, especially in scenarios where smaller models may struggle.

\section{Evaluation}

In this chapter, we provide a comprehensive evaluation of the finetuned models. Our analysis is twofold: firstly, we assess the performance of various model configurations on the semi-synthetic dataset. This evaluation will delve into how different configurations perform in terms of efficacy, which will be measured using a range of metrics.

Secondly, we extend our evaluation to include an analysis of crawled web content. This is a vital step towards demonstrating the real-world applicability of our models. 

\subsection{Evaluation Metrics}
\label{evaluation_metrics}

For automatic evaluation, we apply three rule-based metrics commonly used to evaluate simplification models. Each metric compares reference simplifications with model predictions mostly based on n-gram overlaps. N-grams are contiguous sequences of words in a text.

\textit{BLEU} \cite{papineni_bleu_2002} computes precision scores that measure the frequency of distinct n-grams in the reference simplification over the frequency of distinct n-grams in the model prediction. Typically, we use precision scores for uni, bi, tri, and tetra-grams. These are aggregated with a geometric mean and combined with a brevity penalty for too short predictions.

\textit{METEOR} \cite{banerjee_meteor_2005} is based on matching unigrams of the model's prediction with unigrams of the reference. It calculates precision and a heavily weighted recall on these matches. Additionally, it includes a fragmentation penalty that penalizes predictions with limited sequential overlap with the reference.

\textit{SARI} \cite{xu_optimizing_2016} is designed to evaluate sentence-level text simplification systems. The metric compares n-gram operations between input on the one side and reference and predicted output on the other. It computes F-scores over added and kept n-grams. For deleted n-grams, the precision score is considered. The final score is the arithmetic mean.

\subsection{Automatic Evaluation Results}

\begin{table}[ht]
\caption{Rule-based evaluation metrics computed on the test set. Scores are grouped by pretrained language model and generation algorithm. Metrics increase with model size. For the largest model, beam search is the best decoding algorithm.} 
\centering
\scriptsize 
\begin{tabular}{p{.7cm}ccc}
\toprule
\addlinespace[0.5ex]
& BLEU & METEOR & SARI \\
\midrule
\multicolumn{4}{l}{\textit{Model: gpt2-wechsel-german}} \\
\addlinespace[0.25ex]
greedy  & $0.72 $         & $9.14 $           & $36.61 $  \\
beam    & $\mathbf{1.49}$ & $\mathbf{13.03} $ & $36.80 $  \\
sampling  & $0.96 $       & $11.31 $          & $\mathbf{37.62}$  \\
contrastive  & $0.83 $    & $10.07 $          & $37.02 $  \\
\addlinespace 
\midrule
\multicolumn{4}{l}{\textit{Model: gpt2-xl-wechsel-german}} \\
\addlinespace[0.25ex]
greedy  & $6.77 $            & $23.58 $          & $46.49 $  \\
beam    & $8.21 $            & $23.80 $          & $45.41 $  \\
sampling  & $\mathbf{8.35} $ & $\mathbf{26.86} $ & $\mathbf{47.48} $  \\
contrastive  & $6.99 $       & $23.87 $          & $46.74 $  \\
\addlinespace 
\midrule
\multicolumn{4}{l}{\textit{Model: leo-hessianai-7b}} \\
\addlinespace[0.25ex]
greedy       & $24.46 $          & $45.31 $          & $60.51 $  \\
beam         & $\mathbf{25.97} $ & $\mathbf{46.17} $ & $\mathbf{61.35} $  \\
sampling     & $23.79 $          & $44.97 $          & $60.23 $  \\
contrastive  & $24.39 $          & $45.20 $          & $60.45 $  \\
\addlinespace 
\midrule
\multicolumn{4}{l}{\textit{Model: leo-hessianai-13b}} \\
\addlinespace[0.25ex]
greedy  & $24.53 $          & $45.32 $          & $60.52 $  \\
beam    & $\mathbf{25.78} $ & $\mathbf{45.64} $ & $\mathbf{62.24} $  \\
sampling     & $23.93 $          & $45.06 $          & $60.41 $  \\
contrastive  & $24.64 $          & $45.57 $          & $60.66 $  \\
\bottomrule
\end{tabular}
\label{tab:evaluation_results}
\end{table}

The comprehensive results from our automatic evaluation, as detailed in \autoref{tab:evaluation_results}, provide insights into the performance of two variants of pretrained language models across different generation algorithms. In our analysis, all configurations exhibited the highest score for SARI, followed by METEOR and BLEU. This phenomenon is explored in greater detail in subsection \ref{sec:metric_limitations}. 

Metrics increased with an increasing number of model parameters. Notably, the improvement in metrics was evident up to the transition from models with 7 billion to 13 billion parameters, beyond which we observed no significant differences in metrics. We investigate the behavior of these metrics in more detail in subsection \ref{sec:metric_limitations}. However, the superior performance of the 13 billion parameter model in other tasks suggests that the combination of automated metrics and our dataset may not be capable of discerning performance differences \cite{pluester_leolm_2023}. This could potentially lead to erroneous model selection in practical applications.

In many language generation applications, maximization-based methods like beam search are often deemed less suitable due to their propensity for monotonous and repetitive predictions, as opposed to sampling or contrastive search \cite{su_contrastive_2022}. However, our results do not confirm this for our instantiation of text simplification. The findings suggest that the efficacy of these methods may vary depending on the specific nature of the language generation task.

\subsubsection{Limitations of Rule-Based Metrics Employed within Simplification}
\label{sec:metric_limitations}

\begin{figure}[h!]
    \centering
    \includegraphics[scale=0.8]{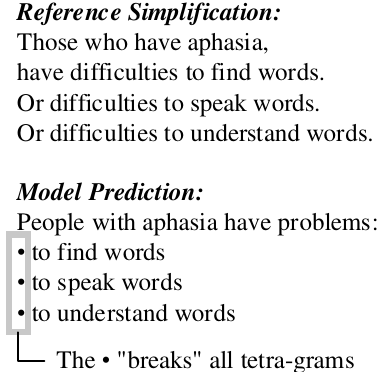}
    \caption{In this example, a reference simplification and a model prediction, translated into English, are contextually similar but lack any shared tetra-grams, yielding a BLEU score of zero.}
    \label{fig:bleu_sample}
\end{figure}

In \autoref{fig:bleu_sample}, different simplification styles stand out. Since our dataset only includes a single reference translation and the metrics focus on sequential overlaps, the stylistic variety of the dataset is not adequately considered in this automated and rule-based evaluation. This leads to lower scores for all metrics we used since all measure the sequential overlap.

\begin{figure}[h!]
    \centering
    \includegraphics[scale=0.325]{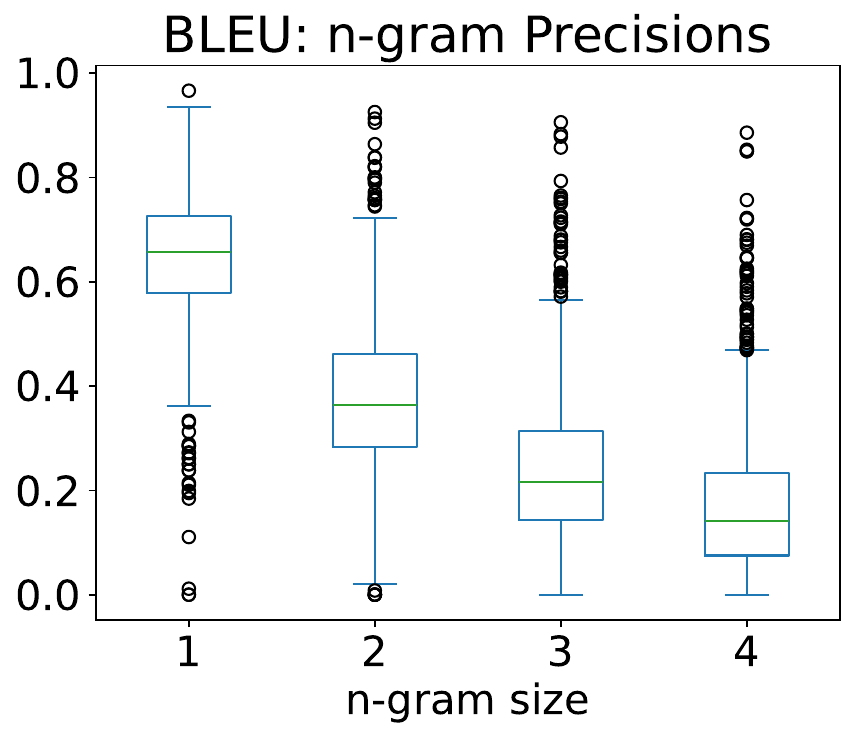}
    \caption{N-gram precisions for predictions of the leo-hessianai-7b model on the complete test set. We observed significantly sloping precision scores for increasing n-gram sizes}
    \label{fig:bleu_ngram_precisions}
\end{figure}

The pair of reference simplification and model prediction in \autoref{fig:bleu_sample} highlights the BLEU metric's key limitations in evaluating text simplification on our test set. The example showcases varying styles between the reference and the model prediction. The reference employs grammatically complete sentences linked with the conjunction "Or", while the model prediction opts for a listing format. This stylistic divergence, especially with short sentences in the model's output, leads to a lack of common 4-grams. BLEU combines n-gram precision with a geometric mean. The geometric mean is calculated by multiplying all the precision values and then taking the $n$th root (where $n$ is the number of values). If any value in the dataset is 0, the product of all the values becomes 0 as well. \autoref{fig:bleu_ngram_precisions} illustrates the n-gram precision scores. This leads to a BLEU score 0, which doesn't accurately reflect the simplified text's quality. By looking at the graph of n-gram precision scores, we can apply this understanding to most of our dataset.



\begin{figure}[h!]
    \centering
    \includegraphics[scale=0.8]{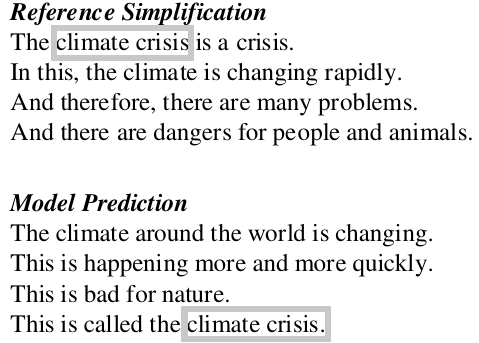}
    \caption{Example to illustrate a high fragmentation penalty due to varied placement of 'climate crisis', negatively impacting the METEOR Score.}
    \label{meteor_sample}
    \label{fig:meteor_sample}
\end{figure}

In the given example in \autoref{fig:meteor_sample}, the reference simplification introduces the term \textit{climate crisis} in the first sentence, whereas the model's simplification describes aspects of the climate crisis before introducing the term. Such rearrangements lead to cross-alignments and higher fragmentation penalties, which affect the METEOR score. This illustrates how n-gram intersections influence the metrics, particularly in the context of differing simplification styles within our corpus. METEOR caps the fragmentation penalty at 50 percent, which limits its influence on the final score.

As indicated in our analysis and shown in \autoref{tab:evaluation_results}, the SARI score tends to rate the model solutions more favorably, aligning more closely with the positive manual evaluations of the models in \autoref{tab:real_evaluation}. This suggests that SARI might be a more reliable indicator of text simplification quality in our context.

SARI was originally meant to be applied within sentence-level simplification. We apply SARI to our multi-sentence documents and we conducted further investigations on the composition of the SARI scores to conclude the metric's plausibility in its three categories, \textit{add}, \textit{keep}, and \textit{delete} within our task. 

In sentence-level text simplification, those operations are considered equally difficult \cite{xu_optimizing_2016}, and therefore, they are weighted equally in the final arithmetic mean. However, due to stylistic transformations within simplification, n-grams of the input are deleted at a much higher probability. For example, structural unigrams like bullet points and line breaks discontinuing original n-grams.

\begin{figure}[h!]
    \centering
    \includegraphics[scale=0.325]{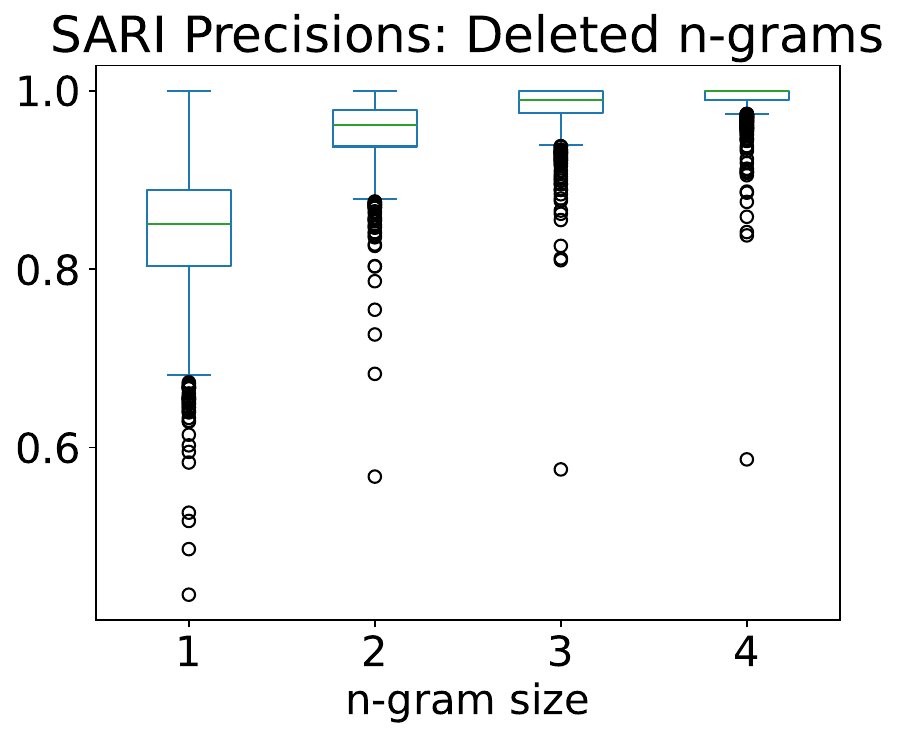}
    \caption{N-gram deletion precisions for predictions of the leo-hessianai-7b model on the complete test set. The median values of our observed SARI delete precision scores reach high values, especially for tri- and tetra-grams}
    \label{fig:sari_delete_precisions}
\end{figure}

Most n-grams from the input, especially tri-grams and tetra-grams, are considered as \textit{correctly deleted} for reference simplification and model prediction. This results in a precision score for deleted tri-grams and tetra-grams close to one for most samples, as illustrated in \autoref{fig:sari_delete_precisions}. Due to the arithmetic mean, this has a nearly constant and strong influence on the final value. SARI tends to be biased optimistically within our task.

Concerning the automated metrics, the 13 billion parameter model does not outperform the 7 billion one. One reason might be the stylistic diversity in the dataset. Given varied styles within even single domains, see \autoref{tab:ndr_mdr_nachrichten_leicht_compare}, the model simplification and the reference simplification might be in different styles. This random factor may affect all metrics that measure sequential overlaps, limiting the overall scores. We might not be able to measure strong models abilities on the target task.




\subsection{Evaluation on Real-World Data}
\label{sec:real_word_evaluation}
We aimed to prove the practical relevance of our models by expanding our evaluation to include real-world data. As evaluation data, we use texts from a German news website\footnote{\url{tagesschau.de}}, a sports news wesite\footnote{\url{transfermarkt.de}}, and a website for tabloid news\footnote{\url{vip.de}}. These texts are not simplified. 

We do not have reference simplifications for these texts. To evaluate the models' simplifications, we consider two types of metrics. Firstly, linguistic simplification using the metrics already introduced. Secondly, the content similarity is done through a manual evaluation of 135 pairs of crawled and non-simplified texts with the simplifications of the models. In this process, pairs could be rated with 0 (no agreement), 1 (partial agreement), 2 (substantial agreement), and 3 (complete agreement). The results in \autoref{tab:real_evaluation} measure the language simplification capabilities of gpt2-xl-wechsel-german (gpt2-xl) and leo-hessianai-7b (leo-7b).

\begin{table}[ht]
\caption{Language complexity and content similarity metrics for model simplifications of real-world online data. \textit{Human Evaluation} summarizes a manual evaluation of content similarity with scores from 0 (no similarity) to 3 (complete equality).}
\centering
\scriptsize 
\begin{tabular}{p{1.35cm}cc}
\toprule
\addlinespace[0.5ex]
\textit{Metrik} & gpt2-xl & leo-7b \\
\midrule
\multirow{2}{=}{\textit{Sentence length}} & \multirow{2}{*}{$16.35 \pm 6.05$} & \multirow{2}{*}{$14.03 \pm 3.47$} \\
& & \\
\midrule
\multirow{2}{=}{\textit{Commas per sentence}} & \multirow{2}{*}{$.48 \pm .68$} & \multirow{2}{*}{$.24 \pm .28$} \\
& & \\
\midrule
\multirow{2}{=}{\textit{Words per line}} & \multirow{2}{*}{$12.35 \pm 6.08$} & \multirow{2}{*}{$10.35 \pm 3.48$} \\
& & \\
\midrule
\multirow{2}{=}{\textbf{\textit{Human Evaluation}}} & \multirow{2}{*}{$1.34 \pm 1.11$} & \multirow{2}{*}{$2.68 \pm 0.55$} \\
& & \\
\bottomrule
\end{tabular}
\label{tab:real_evaluation}
\end{table}

The complexity metrics sentence length, commas per sentence, and words per line indicate that gpt2-xl simplifies texts less than leo-7b. Regarding content accuracy, leo-7b outperformed gpt2-xl, demonstrating a more consistent replication of original content, as shown by the human evaluation scores.

Compared to the gpt2-xl model, the leo-7b model reproduces content much more accurately. On average, the content agreement of this model's simplifications and the original text was rated at least as "substantial agreement".

These results on real data suggest that our models, trained on semi-synthetic data, significantly reduce text complexity while reliably retaining content. Semi-synthetic data is a promising way to train text simplification models and circumvent data scarcity problems.

\section{Limitations}

Our examination reveals that rule-based metrics have limited suitability for evaluating state-of-the-art models in document-level simplification. While our chosen evaluation methodology yields promising results, it lacks a targeted analysis of the end-users for whom the simplification is intended, a scope beyond the ambit of this study. Furthermore, alternative methods to simplify language using LLMs, such as in few-shot learning, merit a comparative analysis against our approach.

\section{Conclusion}

This study represents a significant stride in tackling the challenge of data scarcity in automatic text simplification. We have crafted a semi-synthetic dataset that has proven effective for training models, which are capable of simplifying complex texts. Notably, our models trained on this synthetic data have demonstrated proficiency in simplifying real web content, validating the practicality of our approach. Semi-synthetic data offers the opportunity to efficiently integrate large amounts of existing simplifications into supervised training without manual effort. This is an efficient and promising alternative to alignments or the manual creation of parallel data.

A vital contribution of this work is the open availability of both the dataset and the models, which serve as a foundational resource for researchers in the text simplification field. Integrating state-of-the-art LLMs with supervised learning has shown to be an efficient method for German text simplification. The limitations of current automated and rule-based metrics, such as BLEU, METEOR, and SARI, are increasingly apparent, particularly for our document-level simplification dataset. This suggests that state-of-the-art LLMs may advance to a point where more nuanced evaluation methodologies are required to accurately measure performance differences and select superior models.

Looking ahead, there are promising directions for future research. One crucial area is adapting generation to adhere more closely to the specific simplification styles, for example, following the rules of easy language and thereby tailoring simplifications more effectively to target audiences. This could involve exploring ways to influence the style of simplification, which would enhance the applicability of these models in real-world applications. By fine-tuning the models to align with the nuanced requirements of different user groups, we can make significant strides toward more inclusive and accessible digital content.

\section*{Acknowledgments}
This project is funded by the German Federal Ministry for Family Affairs, Senior Citizens, Women and Youth under the grant number 3923406K05.

\bibliography{ErLeSen}

\begin{thebibliography}{28}
\expandafter\ifx\csname natexlab\endcsname\relax\def\natexlab#1{#1}\fi

\bibitem[{Alva-Manchego et~al.(2021)Alva-Manchego, Scarton, and Specia}]{alva-manchego_suitability_2021}
Fernando Alva-Manchego, Carolina Scarton, and Lucia Specia. 2021.
\newblock \href {https://doi.org/10.1162/coli_a_00418} {The ({Un}){Suitability} of {Automatic} {Evaluation} {Metrics} for {Text} {Simplification}}.
\newblock \emph{Computational Linguistics}, 47(4):861--889.
\newblock Place: Cambridge, MA Publisher: MIT Press.

\bibitem[{Anschütz et~al.(2023)Anschütz, Oehms, Wimmer, Jezierski, and Groh}]{anschutz_language_2023}
Miriam Anschütz, Joshua Oehms, Thomas Wimmer, Bartłomiej Jezierski, and Georg Groh. 2023.
\newblock \href {https://doi.org/10.48550/ARXIV.2305.12908} {Language {Models} for {German} {Text} {Simplification}: {Overcoming} {Parallel} {Data} {Scarcity} through {Style}-specific {Pre}-training}.

\bibitem[{Aumiller and Gertz(2022)}]{aumiller_klexikon_2022}
Dennis Aumiller and Michael Gertz. 2022.
\newblock \href {https://www.semanticscholar.org/paper/Klexikon%3A-A-German-Dataset-for-Joint-Summarization-Aumiller-Gertz/9a89adc19bc33b02afc7ce5e9cab0c690e09fd4b} {Klexikon: {A} {German} {Dataset} for {Joint} {Summarization} and {Simplification}}.

\bibitem[{Banerjee and Lavie(2005)}]{banerjee_meteor_2005}
Satanjeev Banerjee and Alon Lavie. 2005.
\newblock \href {https://aclanthology.org/W05-0909} {{METEOR}: An automatic metric for {MT} evaluation with improved correlation with human judgments}.
\newblock In \emph{Proceedings of the {ACL} Workshop on Intrinsic and Extrinsic Evaluation Measures for Machine Translation and/or Summarization}, pages 65--72, Ann Arbor, Michigan. Association for Computational Linguistics.

\bibitem[{Coster and Kauchak(2011)}]{coster_simple_2011}
W.~Coster and David Kauchak. 2011.
\newblock \href {https://www.semanticscholar.org/paper/Simple-English-Wikipedia%3A-A-New-Text-Simplification-Coster-Kauchak/40c0bab2595a146a5a46c89014172ced20b86866} {Simple {English} {Wikipedia}: {A} {New} {Text} {Simplification} {Task}}.

\bibitem[{Deilen et~al.(2023)Deilen, Garrido, Lapshinova-Koltunski, and Maaß}]{deilen_using_2023}
Silvana Deilen, Sergio~Hernández Garrido, Ekaterina Lapshinova-Koltunski, and Christiane Maaß. 2023.
\newblock \href {https://doi.org/10.48550/ARXIV.2308.11563} {Using {ChatGPT} as a {CAT} tool in {Easy} {Language} translation}.

\bibitem[{Ebling et~al.(2022)Ebling, Battisti, Kostrzewa, Pfütze, Rios, Säuberli, and Spring}]{ebling_automatic_2022}
Sarah Ebling, Alessia Battisti, Marek Kostrzewa, Dominik Pfütze, Annette Rios, Andreas Säuberli, and Nicolas Spring. 2022.
\newblock \href {https://doi.org/10.3389/fcomm.2022.706718} {Automatic {Text} {Simplification} for {German}}.
\newblock In \emph{Frontiers in {Communication}}, volume~7, page 706718.
\newblock ISSN: 2297-900X Journal Abbreviation: Front. Commun.

\bibitem[{Garain et~al.(2019)Garain, Basu, Dawn, and Naskar}]{garain_sentence_2019}
Avishek Garain, Arpan Basu, Rudrajit Dawn, and Sudip~Kumar Naskar. 2019.
\newblock \href {https://doi.org/10.1109/ISCON47742.2019.9036207} {Sentence {Simplification} using {Syntactic} {Parse} trees}.
\newblock \emph{2019 4th International Conference on Information Systems and Computer Networks (ISCON)}, pages 672--676.

\bibitem[{Hauser et~al.(2022)Hauser, Vamvas, Ebling, and Volk}]{hauser_multilingual_2022}
Renate Hauser, Jannis Vamvas, Sarah Ebling, and M.~Volk. 2022.
\newblock \href {https://www.semanticscholar.org/paper/A-Multilingual-Simplified-Language-News-Corpus-Hauser-Vamvas/6c18a5c36bcdf740706c007f4b14a10b8dbc0b1a} {A {Multilingual} {Simplified} {Language} {News} {Corpus}}.

\bibitem[{Hewett(2023)}]{hewett_apa-rst_2023}
Freya Hewett. 2023.
\newblock \href {https://doi.org/10.18653/v1/2023.codi-1.23} {{APA}-{RST}: {A} {Text} {Simplification} {Corpus} with {RST} {Annotations}}.
\newblock \emph{Proceedings of the 4th Workshop on Computational Approaches to Discourse (CODI 2023)}, pages 173--179.

\bibitem[{Holmer and Rennes(2023)}]{holmer_constructing_2023}
Daniel Holmer and Evelina Rennes. 2023.
\newblock \href {https://www.semanticscholar.org/paper/Constructing-Pseudo-parallel-Swedish-Sentence-for-Holmer-Rennes/2b806d60660fd224c2c2601a3c542591ba589423} {Constructing {Pseudo}-parallel {Swedish} {Sentence} {Corpora} for {Automatic} {Text} {Simplification}}.

\bibitem[{Holtzman et~al.(2020)Holtzman, Buys, Du, Forbes, and Choi}]{holtzman_curious_2020}
Ari Holtzman, Jan Buys, Li~Du, Maxwell Forbes, and Yejin Choi. 2020.
\newblock \href {https://doi.org/10.48550/arXiv.1904.09751} {The {Curious} {Case} of {Neural} {Text} {Degeneration}}.
\newblock ArXiv:1904.09751 [cs].

\bibitem[{Jain et~al.(2020)Jain, Patel, Nagalapatti, Gupta, Mehta, Guttula, Mujumdar, Afzal, Sharma~Mittal, and Munigala}]{jain_overview_2020}
Abhinav Jain, Hima Patel, Lokesh Nagalapatti, Nitin Gupta, Sameep Mehta, Shanmukha Guttula, Shashank Mujumdar, Shazia Afzal, Ruhi Sharma~Mittal, and Vitobha Munigala. 2020.
\newblock \href {https://doi.org/10.1145/3394486.3406477} {Overview and {Importance} of {Data} {Quality} for {Machine} {Learning} {Tasks}}.
\newblock In \emph{Proceedings of the 26th {ACM} {SIGKDD} {International} {Conference} on {Knowledge} {Discovery} \& {Data} {Mining}}, {KDD} '20, pages 3561--3562, New York, NY, USA. Association for Computing Machinery.

\bibitem[{Kajiwara and Komachi(2018)}]{kajiwara_text_2018}
Tomoyuki Kajiwara and Mamoru Komachi. 2018.
\newblock \href {https://doi.org/10.5715/jnlp.25.223} {Text {Simplification} without {Simplified} {Corpora}}.
\newblock In \emph{Journal of {Natural} {Language} {Processing}}, volume~25, pages 223--249.
\newblock ISSN: 1340-7619, 2185-8314 Issue: 2 Journal Abbreviation: Journal of Natural Language Processing.

\bibitem[{Maddela et~al.(2022)Maddela, Dou, Heineman, and Xu}]{maddela_lens_2022}
Mounica Maddela, Yao Dou, David Heineman, and Wei Xu. 2022.
\newblock \href {https://doi.org/10.48550/ARXIV.2212.09739} {{LENS}: {A} {Learnable} {Evaluation} {Metric} for {Text} {Simplification}}.

\bibitem[{Minixhofer et~al.(2022)Minixhofer, Paischer, and Rekabsaz}]{minixhofer_wechsel_2022}
Benjamin Minixhofer, Fabian Paischer, and Navid Rekabsaz. 2022.
\newblock \href {https://doi.org/10.18653/v1/2022.naacl-main.293} {{WECHSEL}: {Effective} initialization of subword embeddings for cross-lingual transfer of monolingual language models}.
\newblock In \emph{Proceedings of the 2022 {Conference} of the {North} {American} {Chapter} of the {Association} for {Computational} {Linguistics}: {Human} {Language} {Technologies}}, pages 3992--4006, Seattle, United States. Association for Computational Linguistics.

\bibitem[{Netzwerk-Leichte-Sprache(2022)}]{netzwerk_leichte_sprache_regeln_2024}
Netzwerk-Leichte-Sprache. 2022.
\newblock \href {https://www.leichte-sprache.org/leichte-sprache/die-regeln/} {Die {Regeln} - {Netzwerk} {Leichte} {Sprache}}.

\bibitem[{OpenAI(2023)}]{openai_gpt-4_2023}
OpenAI. 2023.
\newblock \href {https://doi.org/10.48550/arXiv.2303.08774} {{GPT}-4 {Technical} {Report}}.
\newblock ArXiv:2303.08774 [cs].

\bibitem[{Papineni et~al.(2002)Papineni, Roukos, Ward, and Zhu}]{papineni_bleu_2002}
Kishore Papineni, Salim Roukos, Todd Ward, and Wei-Jing Zhu. 2002.
\newblock \href {https://doi.org/10.3115/1073083.1073135} {Bleu: a {Method} for {Automatic} {Evaluation} of {Machine} {Translation}}.
\newblock In \emph{Proceedings of the 40th {Annual} {Meeting} of the {Association} for {Computational} {Linguistics}}, pages 311--318, Philadelphia, Pennsylvania, USA. Association for Computational Linguistics.

\bibitem[{Plüster(2023)}]{pluester_leolm_2023}
Björn Plüster. 2023.
\newblock \href {https://laion.ai/blog/leo-lm} {{LeoLM}: {Igniting} {German}-{Language} {LLM} {Research} {\textbar} {LAION}}.

\bibitem[{Praveen~Kumar et~al.(2022)Praveen~Kumar, Nayak, Shenoy~K., Manoj, and Priyadarshi}]{praveen_kumar_pattern-based_2022}
Archana Praveen~Kumar, Ashalatha Nayak, Manjula Shenoy~K., Roshan~Jacob Manoj, and Akansha Priyadarshi. 2022.
\newblock \href {https://doi.org/10.1109/ACCESS.2022.3174846} {Pattern-{Based} {Syntactic} {Simplification} of {Compound} and {Complex} {Sentences}}.
\newblock \emph{IEEE Access}, 10:53290--53306.

\bibitem[{Radford et~al.(2018)Radford, Narasimhan, Salimans, and Sutskever}]{radford_improving_2018}
Alec Radford, Karthik Narasimhan, Tim Salimans, and Ilya Sutskever. 2018.
\newblock Improving {Language} {Understanding} by {Generative} {Pre}-{Training}.

\bibitem[{Rios et~al.(2021)Rios, Spring, Kew, Kostrzewa, Säuberli, Müller, and Ebling}]{rios_new_2021}
Annette Rios, Nicolas Spring, Tannon Kew, Marek Kostrzewa, Andreas Säuberli, Mathias Müller, and Sarah Ebling. 2021.
\newblock \href {https://doi.org/10.18653/v1/2021.newsum-1.16} {A {New} {Dataset} and {Efficient} {Baselines} for {Document}-level {Text} {Simplification} in {German}}.
\newblock \emph{Proceedings of the Third Workshop on New Frontiers in Summarization}, pages 152--161.

\bibitem[{Spring and Rios(2021)}]{spring_exploring_2021}
Nicolas Spring and Annette Rios. 2021.
\newblock \href {https://doi.org/10.26615/978-954-452-072-4_150} {Exploring {German} {Multi}-{Level} {Text} {Simplification}}.
\newblock In \emph{Proceedings of the {Conference} {Recent} {Advances} in {Natural} {Language} {Processing} - {Deep} {Learning} for {Natural} {Language} {Processing} {Methods} and {Applications}}, pages 1339--1349. INCOMA Ltd. Shoumen, BULGARIA.

\bibitem[{Su and Collier(2022)}]{su_contrastive_2022}
Yixuan Su and Nigel Collier. 2022.
\newblock \href {https://doi.org/10.48550/ARXIV.2210.14140} {Contrastive {Search} {Is} {What} {You} {Need} {For} {Neural} {Text} {Generation}}.
\newblock Publisher: arXiv Version Number: 3.

\bibitem[{Sulem et~al.(2018)Sulem, Abend, and Rappoport}]{sulem_bleu_2018}
Elior Sulem, Omri Abend, and Ari Rappoport. 2018.
\newblock \href {https://doi.org/10.18653/v1/D18-1081} {{BLEU} is {Not} {Suitable} for the {Evaluation} of {Text} {Simplification}}.
\newblock In \emph{Proceedings of the 2018 {Conference} on {Empirical} {Methods} in {Natural} {Language} {Processing}}, pages 738--744, Brussels, Belgium. Association for Computational Linguistics.

\bibitem[{Toborek et~al.(2022)Toborek, Busch, Boßert, Bauckhage, and Welke}]{toborek_new_2022}
Vanessa Toborek, Moritz Busch, Malte Boßert, Christian Bauckhage, and Pascal Welke. 2022.
\newblock \href {https://doi.org/10.48550/ARXIV.2209.01106} {A {New} {Aligned} {Simple} {German} {Corpus}}.
\newblock arXiv.

\bibitem[{Xu et~al.(2016)Xu, Napoles, Pavlick, Chen, and Callison-Burch}]{xu_optimizing_2016}
Wei Xu, Courtney Napoles, Ellie Pavlick, Quanze Chen, and Chris Callison-Burch. 2016.
\newblock \href {https://doi.org/10.1162/tacl_a_00107} {Optimizing {Statistical} {Machine} {Translation} for {Text} {Simplification}}.
\newblock \emph{Transactions of the Association for Computational Linguistics}, 4:401--415.

\end{thebibliography}

\appendix

\section{Example Appendix}
\label{sec:appendix}
\scriptsize
\begin{table}[ht]
\caption{Parameter settings for various algorithms}
\centering
\begin{tabular}{ll}
\toprule
Parameter & Wert \\
\midrule
\multicolumn{2}{l}{\textit{Finetuning}} \\
learning\_rate & 2e-5 \\
weight\_decay & 0.05 \\
batch\_size & 2 \\
n\_epochs & 3 \\
\midrule
\multicolumn{2}{l}{\textit{Greedy}} \\
no\_ngram\_repeat\_size & 5 \\
max\_length & 1024 \\
\midrule
\multicolumn{2}{l}{\textit{Beam Search}} \\
no\_ngram\_repeat\_size & 5 \\
max\_length & 1024 \\
num\_beams & 5 \\
early\_stopping & True \\
\midrule
\multicolumn{2}{l}{\textit{Sampling}} \\
no\_ngram\_repeat\_size & 5 \\
max\_length & 1024 \\
do\_sample & True \\
top\_p & 0.95 \\
top\_k & 5 \\
temperature & 0.5 \\
\midrule
\multicolumn{2}{l}{\textit{Contrastive}} \\
no\_ngram\_repeat\_size & 5 \\
max\_length & 1024 \\
penalty\_alpha & 0.05 \\
top\_k & 5 \\
\bottomrule
\end{tabular}
\label{tab:algorithm_parameters}
\end{table}

\end{document}